\documentclass{article}

\PassOptionsToPackage{numbers, compress}{natbib}

\usepackage[final]{neurips_2022}

\usepackage[utf8]{inputenc} 
\usepackage[T1]{fontenc}    
\usepackage[pagebackref=true,breaklinks=true,colorlinks,citecolor=citecolor, linkcolor=linkcolor]{hyperref}       
\usepackage{url}            
\usepackage{booktabs}       
\usepackage{amsfonts}       
\usepackage{nicefrac}       
\usepackage{microtype}      
\usepackage[table]{xcolor}  
\usepackage{bm}

\definecolor{citecolor}{HTML}{0071BC}
\definecolor{linkcolor}{HTML}{ED1C24}

\title{AdaptFormer: Adapting Vision Transformers for Scalable Visual Recognition}

\author{%
  \vspace{0.4em}
  \centerline{
  Shoufa Chen$^{1}$\thanks{Equal contribution.} \quad
  Chongjian Ge$^{1*}$ \quad
  Zhan Tong$^{2}$ \quad
  Jiangliu Wang$^{2}$ } \\
  \textbf{
  Yibing Song$^{2}$ \quad
  Jue Wang$^{2}$ \quad
  Ping Luo$^{1}$} \vspace{0.6em}
  \\
  \centerline{
  $^{1}$The University of Hong Kong   \quad \quad
  $^{2}$Tencent AI Lab 
  }
}

\usepackage{multirow}
\usepackage{booktabs}
\usepackage{subcaption}
\usepackage{amsmath,amsfonts,bm,amsthm,amssymb}
\usepackage{algorithm}
\usepackage{algorithmic}
\usepackage{graphics,graphicx,caption,float,color}
\usepackage{wrapfig}
\usepackage{bbm}
\usepackage{appendix}

\usepackage{listings}
\usepackage{xspace}
 
\usepackage{hhline}
\usepackage{multicol}
\usepackage{booktabs} 
\usepackage{makecell} 

\newcommand{\ourabbr}{AdaptFormer\xspace}

\makeatletter
\newcommand*\bigcdot{\mathpalette\bigcdot@{.5}}
\newcommand*\bigcdot@[2]{\mathbin{\vcenter{\hbox{\scalebox{#2}{$\m@th#1\bullet$}}}}}
\makeatother

\definecolor{MyDarkBlue}{rgb}{0,0.08,1}
\definecolor{MyDarkGreen}{rgb}{0.02,0.6,0.02}
\definecolor{MyDarkRed}{rgb}{0.8,0.02,0.02}
\definecolor{MyDarkOrange}{rgb}{0.40,0.2,0.02}
\definecolor{MyPurple}{RGB}{111,0,255}
\definecolor{MyRed}{rgb}{1.0,0.0,0.0}
\definecolor{MyGold}{rgb}{0.75,0.6,0.12}
\definecolor{MyDarkgray}{rgb}{0.66, 0.66, 0.66}

\definecolor{Image}{RGB}{93,59,155}
\definecolor{Video}{RGB}{105,145,60}

\definecolor{ice}{RGB}{84,146,214}
\definecolor{fire}{RGB}{217,129,63}
\definecolor{baselinecolor}{gray}{.9}
\definecolor{poscolor}{RGB}{212,17,89}
\definecolor{negcolor}{RGB}{26,133,255}
\definecolor{bestcolor}{RGB}{238, 255, 238}
\newcolumntype{x}[1]{>{\centering\arraybackslash}p{#1pt}}
\newcolumntype{y}[1]{>{\raggedright\arraybackslash}p{#1pt}}
\newcolumntype{z}[1]{>{\raggedleft\arraybackslash}p{#1pt}}
\newcommand{\tablestyle}[2]{\setlength{\tabcolsep}{#1}\renewcommand{\arraystretch}{#2}\centering}

\newcommand{\baseline}[1]{\cellcolor{baselinecolor}{#1}}
\newcommand{\bestcell}[1]{\cellcolor{bestcolor}{#1}}
\newcommand{\aaa}[1]{\textcolor{poscolor}{\small (#1)}}
\newcommand{\bbb}[1]{\textcolor{negcolor}{\small (#1)}}

\newcommand{\rebuttal}[1]{#1}

\newcommand{\eg}{\emph{e.g.}}
\newcommand{\ie}{\emph{i.e.}}
\newcommand{\etal}{\emph{et.al.,}}

\begin{document}

\maketitle

\begin{abstract}

Pretraining Vision Transformers (ViTs) has achieved great success in visual recognition. A following scenario is to adapt a ViT to various image and video recognition tasks. The adaptation is challenging because of heavy computation and memory storage. Each model needs an independent and complete finetuning process to adapt to different tasks, which limits its transferability to different visual domains.
To address this challenge, we propose an effective adaptation approach for Transformer, namely \ourabbr, which can adapt the pre-trained ViTs into many different image and video tasks efficiently.
It possesses several benefits more appealing than prior arts.
Firstly, \ourabbr introduces lightweight modules that only add less than 2\% extra parameters to a ViT, while it is able to increase the ViT's transferability without updating its original pre-trained parameters, significantly outperforming the existing 100\% fully fine-tuned models on action recognition benchmarks.
Secondly, it can be plug-and-play in different Transformers and scalable to many visual tasks.
Thirdly, extensive experiments on five image and video datasets show that \ourabbr largely improves ViTs in the target domains. For example, when updating just 1.5\% extra parameters, it achieves about 10\% and 19\% relative improvement compared to the fully fine-tuned models on Something-Something~v2 and HMDB51, respectively. 
Code is available at \url{https://github.com/ShoufaChen/AdaptFormer}.

\end{abstract}
\section{Introduction}

There is a growing interest in adopting a general neural model to tackle a large variety of different tasks since it benefits in reducing the need for task-specific model design and training. Recently, Transformer~\cite{vaswani-nips17-transformer} demonstrates great potential in this goal considering its success in various fields, \eg, natural language processing~(NLP)~\cite{devlin-2018-bert, brown-nips20-gpt, wang2018glue, yang2019xlnet}, visual recognition~\cite{dosovitskiy-2020-vit, touvron2020training, yuan2021tokens, liu-iccv21-swin}, dense prediction~\cite{wang2021pyramid, carion-eccv20-detr, zhu-2020-deformabledetr, zheng2021rethinking, xie-nips21-segformer}, Generative Adversarial Network~(GAN)~\cite{jiang2021transgan, hudson2021generative}, reinforcement learning~(RL)~\cite{chen2021decision, chen2022transdreamer, yang2022learning}, robotics~\cite{jangir2022look, dasari2020transformers}, and etc. 
However, existing literature in computer vision tend to focus on the \emph{same network with task-specific weights} scenario, where a single network is used to train from scratch or fully fine-tune on a specific dataset, making it infeasible to maintain a separate model weight for every dataset when the number of task grows, especially for the increasing model capacity of state-of-the-art models (\eg, ViT-G/14~\cite{zhai2021scaling} with over 1.8 billion parameters). 

Different from prior arts, we step into the direction of developing \emph{same network with almost same weights} and achieve superior performance than the full-tuning approach by only tuning less than 2\% parameters, with the remaining over 98\% parameters shared across different tasks. There are two challenges to learning universal representations using a single model. The first one lies in the pre-training stage, which requires algorithms that can learn well-generalized representations that are easy to be applied to many tasks. Recent arts in self-supervised learning~\cite{caron2021emerging, bao-2021-beit, he-2021-mae, zhou2021ibot, wei2021masked, tong-2022-videomae, feichtenhofer2022masked} can serve as a solution to this challenge. The second one, which is our main concern in this work, is to build an effective pipeline that can adapt the model obtained at the pre-training stage to various downstream tasks by tuning parameters as less as possible and keeping the left parameters frozen. 

While fine-tuning pre-trained models has been widely studied in NLP~\cite{bapna2019simple, houlsby-2019-adapternlp, pfeiffer-2020-adapterfusion, pfeiffer-2020-adapterhub, li-2021-prefix, lester-2021-nlpprompt, hu-2021-lora, zaken-2021-bitfit, liu-2021-ptuningv2, he2022towards}, this topic is seldomly explored in the vision, where full tine-tuning of model parameters is still the dominant strategy for adapting vision transformers. However, the full fine-tuning cannot satisfy the goal of \emph{universal representation} as it assigns an independent set of weights for every task. Linear probing is a straightforward approach to maintaining the pre-trained model fixed by only tuning a specific lightweight classification head for every task. However, linear probing tends to have an unsatisfactory performance and misses the opportunity of pursuing strong but non-linear features~\cite{he-2021-mae},  which indeed benefit deep learning.
\begin{wrapfigure}{R}{0.52\textwidth}
    \vspace{-2.0em}
\centering
\includegraphics[width=0.52\textwidth]{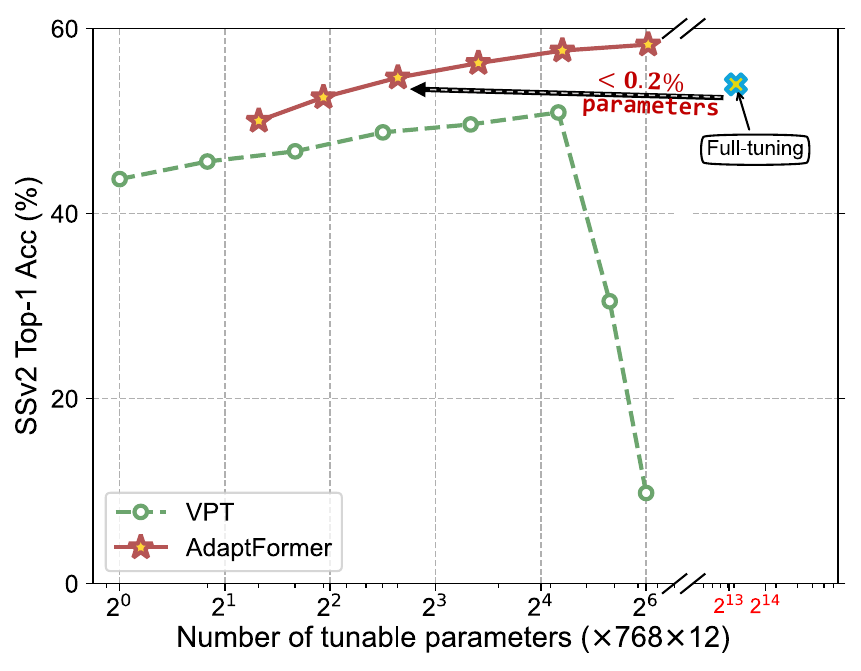}
    \caption{\footnotesize\textbf{Parameter-Accuracy trade-off.} We leverage ViT-Base as backbone and report top-1 accuracy on SSv2 dataset. \ourabbr can surpass full-tuning with only 0.2\% tunable parameters. More detailed results are shown in Table~\ref{tab:ssl_pretrain}.
    }
    \label{fig:intro_result}
    \vspace{-1.0em}
\end{wrapfigure}
More recently, Bahng~{\etal}~\cite{bahng-2022-vp} aimed to adapt pre-trained models by modifying raw input pixel space. Jia~{\etal}~\cite{jia-2022-vpt} proposed Visual Prompt Tuning~(VPT) to adapt transformer models for downstream vision tasks, which prepends several learnable parameters~(\emph{prompts}) to the patch embeddings and freezes the whole pre-trained backbone. 

In this work, we propose a lightweight module, namely \ourabbr, to adapt vision transformers by updating the weights of \ourabbr. We introduce learnable parameters from the model perspective, which is different from VPT, which inserts learnable parameters into the token space. Our \ourabbr is conceptually simple yet effective. It consists of two fully connected layers, a non-linear activation function, and a scaling factor. This module is set in parallel to the feed-forward network~(FFN) of the original ViT model, as shown in Figure~\ref{fig:adaptformer-arch}. This design is turned out to be effective for model transfer when 
processing scalable visual tokens for both image and video data (i.e., image data consists of a small scale of visual tokens while video data consists of a large scale). As shown in Figure~\ref{fig:intro_result}, compared with the full-tuning strategy, AdaptFormer achieves comparable performance on video recognition with only about 0.1\% tunable parameters. Meanwhile, with less than 2\% tunable parameters, AdaptFormer surpasses the full-tuning solution by about 10\% on top-1 accuracy.
Similar approaches are also proposed in fine-tuning pre-trained language models~(PLMs)~\cite{bapna2019simple, houlsby-2019-adapternlp, pfeiffer-2020-adapterhub, he2022towards}.

The key \textbf{contributions} of this paper are summarized as follows:
~\textbf{(1)}~We propose a simple yet effective framework, namely \ourabbr, for adapting vision transformers to a large variety of downstream visual recognition tasks and avoiding catastrophic interference with each other. To the best of our knowledge, this is the first work that explores efficient fine-tuning in video action recognition.
~\textbf{(2)}~We ablate many design choices and demonstrate the superior robustness of \ourabbr when parameters scale up.
~\textbf{(3)}~Extensive experiments on various downstream tasks demonstrate that \ourabbr outperforms existing fine-tuning approaches significantly. 
By demonstrating the effectiveness of \ourabbr on multiple visual benchmarks, we hope our work could inspire the research communities to rethink the fine-tuning mechanism in computer vision and make progress toward a flexible yet universal Transformer model for visual recognition.

\vspace{-2mm}
\section{Related Works}
\vspace{-2mm}

In the proposed \ourabbr, we mainly introduce a plug-and-play module for efficiently fine-tuning the current vision Transformer models. In this section, we perform a literature review on related works from two perspectives, \ie, the vision Transformers, and efficient transfer learning for vision Transformers. 

\subsection{Transformer in Vision}
The Transformer architecture is first introduced in~\cite{vaswani-nips17-transformer} and has re-energized the natural language processing (NLP) field from then on~\cite{devlin-2018-bert,brown-nips20-gpt}. Inspired by its huge success, researches in the computer vision filed have also evolved into Transformer era since ViTs~\cite{dosovitskiy-2020-vit}. The strong capability of modeling long-range relation has facilitated Transformer in various vision tasks, including image classification~\cite{dosovitskiy-2020-vit,liu-iccv21-swin,liang-iclr22-not}, object detection~\cite{carion-eccv20-detr,zhu-2020-deformabledetr,chi-nips20-relationnet++}, semantic/instance segmentation~\cite{xie-nips21-segformer}, video understanding~\cite{bertasius-2021-timesformer, arnab2021vivit, fan-iccv21-mvit, li2022uniformer}, point cloud modeling~\cite{zhao2021point, guo-2021-pct}, 3D Object Recognition~\cite{ chen-2021-mvt} and even low-level processing~\cite{chen-cvpr21-pre,liang-iccv21-swinir,wang-2021-uformer}. Furthermore, transformers have advanced the vision recognition performance by a large-scale pretraining~\cite{chen-iccv21-mocov3,pan-cvpr21-videomoco,caron-iccv21-dino,ge-nips21-care,he-2021-mae,tong-2022-videomae,radford-icml21-clip}. In such 
a situation, given the pre-trained Transformer models, which are more larger than the previously prevalent CNN backbones, one open question is how to fine-tune the big vision models so that they can be adapted into more specific down-stream tasks. To solve the open question, we propose \ourabbr to transfer ViTs from the pre-trained pre-texts into the target tasks in a more effective and efficient way.

\subsection{Efficient Transfer learning for Transformers}
Transfer learning targets re-adopting a pre-trained model (either via the supervised or the unsupervised manner) as the starting point and further fine-tuning the specific model on a new task. In the NLP field, transferring the large pre-trained language models (PLMs)~\cite{devlin-2018-bert,brown-nips20-gpt} into downstream tasks has been the popular paradigm for a long time. Conventional arts~\cite{devlin-2018-bert,brown-nips20-gpt} set all the network parameters as learnable ones and adapt them to the target tasks. However, with the growth of model sizes and the complexity of the specific tasks, the conventional paradigm is inevitably limited by the huge computational burden. The NLP community has explored several ways for parameter-efficient transfer learning that only set a few parameters learnable and fine-tune them for efficiency. The pioneer works could be mainly categorized from the token~\cite{li-2021-prefix,lester-2021-nlpprompt} and network perspectives~\cite{houlsby-2019-adapternlp,hu-2021-lora,zaken-2021-bitfit,guo-2020-diff}. 
Basically speaking, the token-related methods~\cite{lester-2021-nlpprompt,li-2021-prefix} typically prepend several learnable preﬁx vectors/tokens to the projected tokens within the multi-head self-attention layers (MHSA~\cite{vaswani-nips17-transformer}). The philosophy behind it is to assist the pre-trained models in understanding downstream tasks with the guidance of extra token information. On the other hand, 
network-related methods~\cite{houlsby-2019-adapternlp,hu-2021-lora} integrate shallow modules to improve the model transferability. The introduced modules adapt the produced representations into the downstream tasks via features fusion.

Recently, with the emergence of a much more large-scale dataset~\cite{deng-cvpr09-imagenet,ridnik-2021-imagenet21k,sun-iccv17-jfm,mahajan-eccv18-instagram,kay-2017-k400}, increasing researchers in computer vision have adopted the homologous paradigm, \ie, first pre-training and then fine-tuning, to advance the vision tasks. As for the second stage, traditional methods typically adopt the full-tuning arts in the downstream tasks. Rare attention has been drawn to the field of efficient adaptation, especially in the field of vision Transformers. Inspired by Prompting in NLP, ~\cite{jia-2022-vpt} introduced the learnable tokens in exploring the efficient adaptation for ViTs. We empirically found that the performance of 
prompting is hindered by the scale of tokens. That is to say, for the tasks where the number of tokens is on a small scale, \eg, image classification, Prompting is efficient for improving the model transferability. However, for larger scale tokens, \eg, video understanding, Prompting presents limited potential.
This observation motivates us to introduce AdaptFormer, which is effective in the scenarios of scalable visual tokens.

\section{Approach}

\begin{figure}[ht]
\centering
    \begin{subfigure}{0.2\textwidth}
    \centering
    \includegraphics[height=1.8\textwidth]{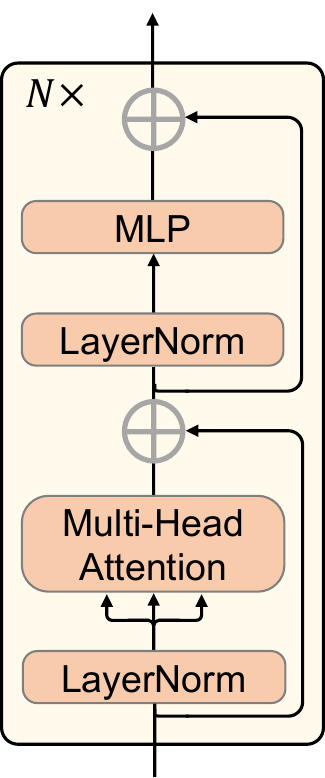}
    \caption{Full fine-tuning.}\label{fig:fulltune-arch}
    \end{subfigure}
    \hspace{0.16\textwidth}
    \begin{subfigure}{0.4\textwidth}
    \centering
    \includegraphics[height=0.9\textwidth]{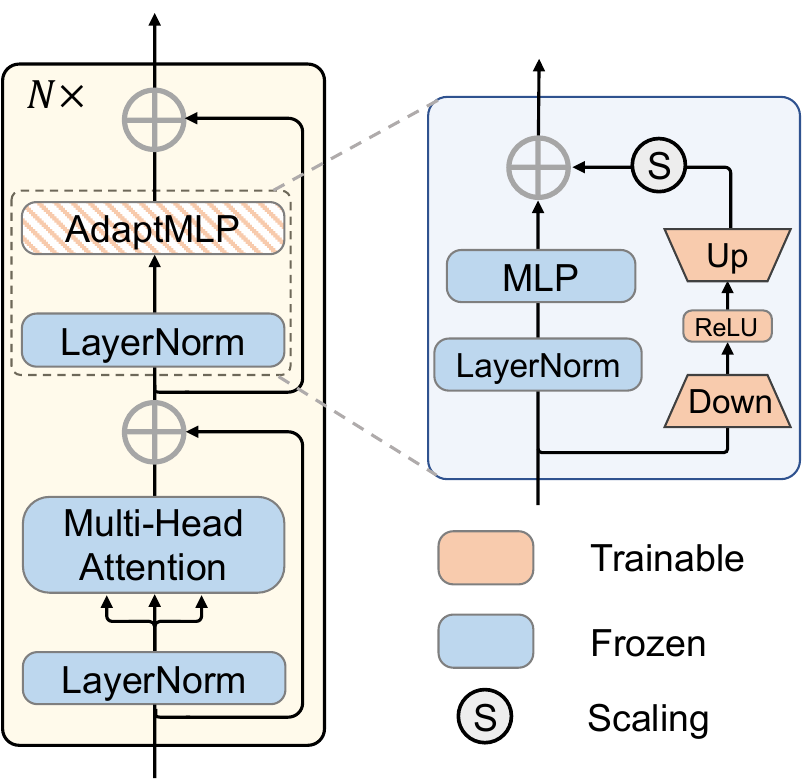}
    \caption{\ourabbr fine-tuning}\label{fig:adaptformer-arch}
    \end{subfigure}
    \caption{\textbf{Comparison of previous \emph{full} and our \emph{\ourabbr} fine-tuning.} \ourabbr is conceptually simple by replacing the original MLP block with AdaptMLP, which consists of two branches, including the frozen branch (left) and the trainable \texttt{down} $\rightarrow$ \texttt{up} bottleneck module (right).}
    \label{fig:tuning-arch}
    \vspace{-1.5em}
\end{figure}

We propose \ourabbr for efficiently transferring large pre-trained vision transformer models to downstream tasks, in both image and video domains. \ourabbr attains strong transfer learning abilities by only fine-tuning a small number of extra parameters, circumventing catastrophic interference among tasks. We illustrate the overall framework of \ourabbr in Figure~\ref{fig:adaptformer-arch}.


\subsection{Preliminary and Notation
}
Vision Transformers (ViTs) are first introduced by~\cite{dosovitskiy-2020-vit} into vision recognition. A vanilla vision Transformer basically consists of a patch embedding layer and several consecutively connected encoders, as depicted in Figure~\ref{fig:fulltune-arch}. Given an image $\mathit{x} \in \mathbb{R}^{H \times W \times 3}$, the patch embedding layer first splits and flatten the sample $\mathit{x}$ into sequential patches $\mathit{x}_p \in \mathbb{R}^{N \times (P^2 d)}$, where $\mathit{(H,W)}$ represents the \emph{height} and \emph{width} of the input image, $(P, P)$ is the resolution of each image patch, $\mathit{d}$ denotes the output channel, and $N = {HW}/{P^2}$ is the number of image tokens.
The overall combination of a prepended \verb|[CLS]| token and the image tokens $\mathit{x}_p$ are further fed into Transformer encoders for attention calculation.

Each Transformer encoder mainly consists of two types of sub-layers, \ie, a multi-head self-attention layer~(MHSA) and a MLP layer. In MHSA, the tokens are linearly projected and further re-formulated into three vectors, namely $\bm{Q}, \bm{K}$ and $\bm{V}$. The self-attention calculation is performed on $\bm{Q}, \bm{K}$ and $\bm{V}$ by:
\vspace{-1pt}
\begin{equation}
    x'_\ell = {\rm Attention}(\bm{Q}, \bm{K}, \bm{V}) = {\rm Softmax}(\frac{\bm{Q} \bm{K}^\top}{\sqrt{d}})\bm{V},
    \label{eq:attn}
\end{equation}
where $x'_\ell$ are the tokens produced by MHSA at the $\ell$-th layer. The output tokens  $x'_\ell$ are further sent to a LayerNorm~\cite{ba-2016-layernorm} and a MLP block which is consisted of two fully connected layers with a GELU activation~\cite{hendrycks-2016-gelu} in between. This process is formally formulated as follows,
\begin{equation}
    \mathit{x}_\ell = {\rm MLP}({\rm LN}(x'_\ell)) + x'_\ell,
    \label{eq:ffn}
\end{equation}
where $x_\ell$ is the output of the $\ell$-th encoder block.
At the last transformer layer, the \verb|[CLS]| is utilized for the final object recognition. We refer the readers to find more details in~\cite{dosovitskiy-2020-vit}. In our work, we replace the MLP layer with our AdaptMLP module for efficient fine-tuning purposes.

\subsection{AdaptFormer}
We propose a plug-and-play bottleneck module, namely AdaptMLP\footnotemark.
\footnotetext{In this paper, we use the term `AdaptMLP' to denote the designed module and the term `AdaptFormer' to represent the fine-tuning framework for Vision Transformers. Unless otherwise specified, we apply AdaptFormer to fine-tune the vanilla ViT backbone~\cite{dosovitskiy-2020-vit} in this paper.}
We denote the vision Transformer equipped with AdaptMLP as AdaptFormer.

\noindent\textbf{Architecture.}
The design principle of \ourabbr is simple yet effective, which is illustrated in Figure~\ref{fig:adaptformer-arch}. Compared to the vanilla full fine-tuning regime, \ourabbr replaces the MLP block in the transformer encoder with \emph{AdaptMLP}, which is consisted of two sub-branches. The MLP layer in the left branch is identical to the original network, while the right branch is an additionally introduced lightweight module for task-specific fine-tuning.
Specifically, the right branch is designed to be a bottleneck structure for limiting the number of parameters purpose, which includes a down-projection layer with parameters $\bm W_{\rm down} \in \mathbb{R}^{d\times\hat{d}}$, an up-projection layer with parameters $\bm W_{\rm up} \in \mathbb{R}^{\hat{d}\times d}$, where $\hat{d}$ is the bottleneck middle dimension and satisfies $\hat{d} \ll d$. In addition, there is a ReLU layer~\cite{agarap-2018-relu} between these projection layers for non-linear property. This bottleneck module is connected to the original MLP network (left branch) through the residual connection via a scale factor $\mathit{s}$.
For a specific input feature $x'_\ell$, the right branch in AdaptMLP produces the adapted features, $\tilde{x}_\ell$, formally via:
\vspace{-1pt}
\begin{equation}
    \tilde{x}_\ell =  {\rm ReLU}({\rm LN}(x'_\ell) \cdot {\bm W_{\rm down}})\cdot {\bm W_{\rm up}}.
    \label{eq:ada_fea}
\end{equation}
Then both the features $ \tilde{x}_\ell$ and $x'_\ell$ are fused with $x_\ell$ by residual connection,
\vspace{-1pt}
\begin{equation}
    x_\ell =  {\rm MLP}({\rm LN}(x'_\ell)) + \mathit{s}\cdot \tilde{x}_\ell + x'_\ell.
    \label{eq:adapter}
\end{equation}

\noindent\textbf{Fine-tuning.} During the fine-tuning phase, we only choose the newly added parameters to optimize and keep rest ones fixed. Specifically, the original model parts~(\textcolor{ice}{blue} blocks in Figure~\ref{fig:adaptformer-arch}) load weights from the pre-trained checkpoint and keeps parameters \textit{frozen}. The newly added parameters (\textcolor{fire}{orange} blocks) are updated on the specific data domain with the task-specific losses. 

\noindent\textbf{Inference.} After fine-tuning, we still keep the shared parameters frozen as in the previous fine-tuning state, and additionally load the weights of the extra parameters that were fine-tuned in the previous stage. The single overall model is able to be adapted to multiple tasks with the assistance of lightweight introduced modules.


\subsection{Discussion}\label{sec:num_parameters}

\noindent{\textbf{Tunable parameters analysis.}}
Our AdaptMLP module is lightweight. The total number of parameters introduced to per layer is $2\times d\times \hat{d} + \hat{d} + d$, which includes biases parameters. The middle dimension $\hat{d}$ is a small value compared with $d$~(\ourabbr still obtains a decent performance even when $\hat{d} = 1$, as discussed in Sec.~\ref{sec:abla}). Since most of the shared parameters are fixed and the number of newly introduced parameters is small~($< 2\%$ of the pre-trained model parameters), the total model size grows slowly when more downstream tasks are added.

\begin{wrapfigure}{R}{0.38\textwidth}
\vspace{-20pt}
\begin{center}
    \includegraphics[width=0.36\textwidth]{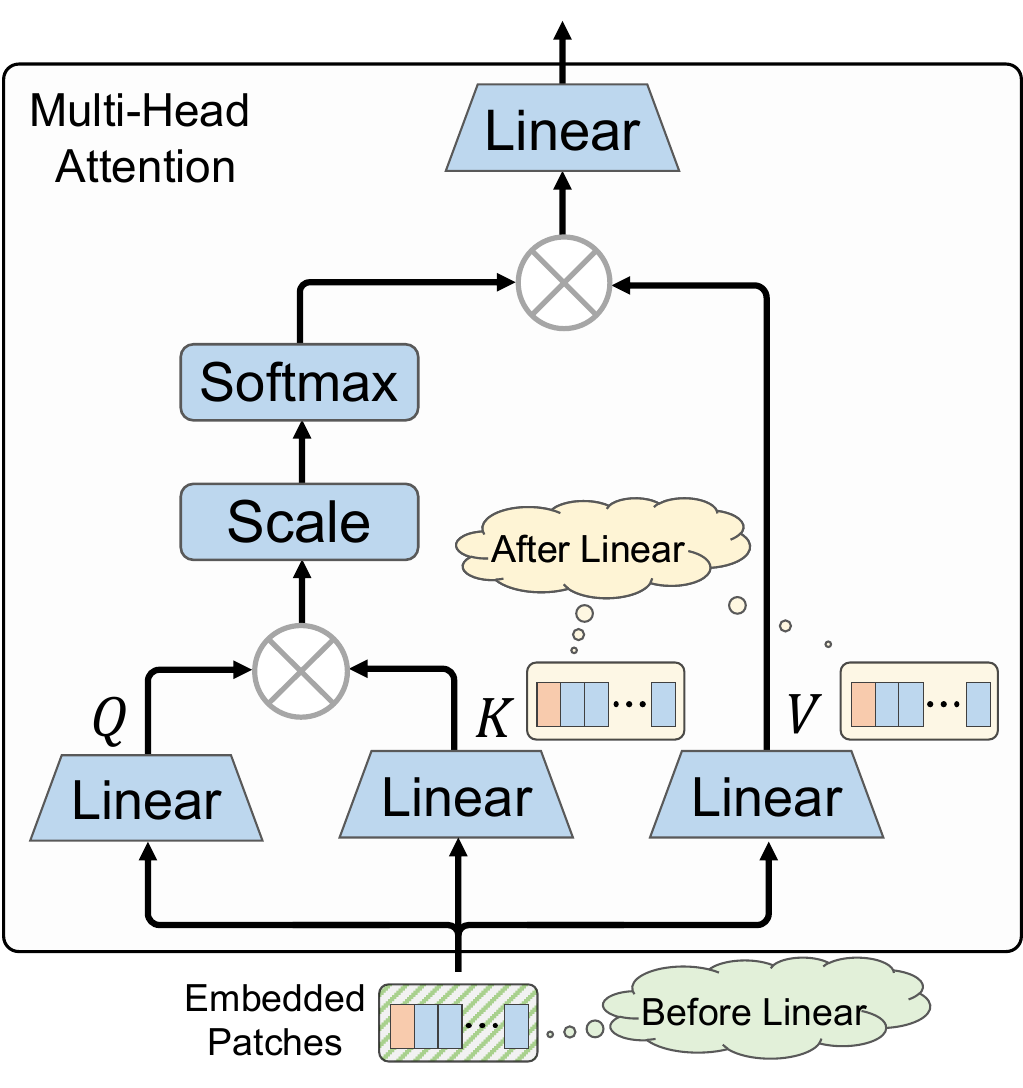}
    \caption{\textbf{Prompt tuning illustration.}}\label{fig:prompting}
\end{center}
\vspace{-10pt}
\end{wrapfigure}

\noindent\textbf{Applicability.} We note that AdaptMLP is a plug-and-play module that can be adaptively inserted into existing popular vision transformer architectures~\cite{dosovitskiy-2020-vit, liu-iccv21-swin, wang2021pyramid, yuan2021tokens, chu2021Twins, dong2021cswin} since all of the backbones share the same MLP layers even though they differ in the MHSA architectures (as shown in Figure~\ref{fig:adaptformer-arch}). Compared to our methods, we notice that recent prompt-related approaches insert trainable parameters into the token space, as illustrated in Figure~\ref{fig:prompting}. They prepend learnable parameters either into the embedded tokens before linear projection~\cite{li-2021-prefix} or the key and value tokens after linear projection~\cite{jia-2022-vpt}. \textit{Therefore, the prompt-related method can not be straightforwardly adapted to special MHSA variants, especially for the one that takes the pyramid spatial information into account}~\cite{liu-iccv21-swin,wang2021pyramid}. Besides,
we empirically observe that prompt-related methods perform not well when the number of patch tokens grows up from image to video scale, as shown in Figure~\ref{fig:intro_result}.

In summary,  we present a strategy for tuning a pre-trained vision Transformer on a set of scalable vision recognition tasks (\eg image domain and video domain). It adds limited learnable parameters for tuning while achieving comparable or even better performance than the full-tuning strategy. Moreover, \ourabbr could serve as a generic module for a large variety of recognition tasks.

\noindent\textbf{Insights of architecture design.}
The MLP module is important for ViTs. As illustrated in~\cite{dong2021attention}, MLPs prevent ViTs from producing a rank-1 matrix. Also, MLPs stop the ViT output from degenerations. Inspired by the above analysis, we believe an effective ViT adaptation shall focus on its MLPs rather than multi-head self attentions. Meanwhile, we learn from the inception framework~\cite{szegedy2015going} that parallel design is an effective way for feature ensemble. With the parallel design, the domain-specific features produced by the adapter module can supplement the domain-agnostic features from the fixed branch for a better feature ensemble. Our following experiments will verify that the parallel performs better than the sequential design.

Besides, though many advanced Transformer-based models~\cite{liu-iccv21-swin, wang2021pyramid, fan2021multiscale, yuan2021tokens} which have emerged since the success of ViT having different attention mechanisms within the Transformer block, they all share the similar MLPs (feed-forward network) structures. Therefore, our AdaptMLP can be easily plugged into these ViT variants. Moreover, AdaptMLP can also be applied to more recent attention-free models~\cite{tolstikhin2021mlp, liu2021pay, chen2022cyclemlp}.

\section{Experiments}

We evaluate the effectiveness of \ourabbr by conducting extensive visual recognition experiments in both the image and video domains. We first describe our experimental settings in Sec.~\ref{Sec:exp_setting}, covering the pre-trained backbones, baseline methods, downstream tasks and training details. We then compare \ourabbr with baseline methods and provide a thorough analysis in Sec.~\ref{sec:main_results}. In addition, we also conduct ablation studies to explore different experimental configurations and explain what makes for the superiority of \ourabbr in Sec~\ref{sec:abla}.

\subsection{Experimental Settings}\label{Sec:exp_setting}
\noindent\textbf{Pre-trained backbone.} We adopt the plain Vision Transformer~(ViT)~\cite{dosovitskiy-2020-vit}, \ie, ViT-Base~(ViT-B/16) as our backbone model and pre-train the model with both supervised and self-supervised approaches.
Specifically, for \textcolor{Image}{\textbf{image}}, we directly use the ImageNet-21k~\cite{deng-cvpr09-imagenet} supervised pre-trained model\footnotemark\footnotetext{\scriptsize\url{https://github.com/rwightman/pytorch-image-models/releases/download/v0.1-vitjx/jx_vit_base_patch16_224_in21k-e5005f0a.pth}}
and MAE~\cite{he-2021-mae} self-supervised model\footnotemark.
For \textcolor{Video}{\textbf{video}}, we take both supervised and self-supervised pre-trained models from VideoMAE~\cite{tong-2022-videomae}. More details about pre-training approaches and datasets can be found in Appendix.

\footnotetext{\scriptsize\url{https://dl.fbaipublicfiles.com/mae/pretrain/mae_pretrain_vit_base.pth}}

\textbf{Initialization of \ourabbr.} For the original networks, we directly load the weights pre-trained on the upstream tasks and keep them frozen/untouched during the fine-tuning process. For the newly added modules, the weights of down-projection layers are initialized with Kaiming Normal~\cite{he2015delving}, while the biases of the additional networks and the weights of the up-projection layers are configured with zero initialization. The reason for the zero initialization of other layers is that in this way, the initial newly added parameters are initialized such that the new function resembles the original one at the start of the fine-tuning stage. We empirically found that if the initialization deviates too far from the identity function, the model is not stable to train.

\noindent\textbf{Baseline methods.} We compare \ourabbr with three commonly used fine-tuning approaches, including (1)\emph{Linear probing:} adding an extra linear layer on top of the backbone and tuning the added parameters for evaluation. (2) \emph{Full Fine-tuning:} setting all the parameters learnable and tuning them together. (3) \emph{Visual Prompt Tuning~(VPT):}~\cite{jia-2022-vpt} fine-tuning the extra token parameters as shown in Figure~\ref{fig:prompting}.

\noindent\textbf{Downstream tasks.}
We evaluate our \ourabbr on both image and video recognition tasks to verify its effectiveness. The specific datasets leveraged in this work are presented in the following.

\noindent\textcolor{Image}{\textbf{$\bullet$\quad Image domain :}}
CIFAR-100~\cite{krizhevsky-2009-cifar} contains 50,000
training images and 10,000 validation images of resolution 32×32 with 100 labels.
Street View House Numbers (SVHN)~\cite{goodfellow-2013-svhn} is a digit classification benchmark dataset. In total, the dataset comprises over 600,000 labeled images, containing 73,257 training samples, 26,032 testing samples and 531,131 extra training data. The Food-101~\cite{bossard-eccv14-food101} dataset consists of 101 food categories with a total of 101k images, including 750 training and 250 testing samples per category.

\noindent\textcolor{Video}{\textbf{$\bullet$\quad Video domain :}}
Something-Something~V2~(SSv2)~\cite{goyal-iccv17-sthsthv2} is a large collection of video clips showing the people perform several normal actions in the daily life (\eg, moving stuff and opening the door). It consists of 168,913 training samples, 24,777 validation samples and 27,157 testing samples, making a total of 220,847 videos with 174 labels. HMDB51~\cite{kuehne-iccv11-hmdb} is composed of 6,849 videos with 51 categories, making a split of 3.5k/1.5k train/val videos.

\noindent\textbf{Implementation details.}
In this work, we use PyTorch toolkit~\cite{NEURIPS2019_9015} to conduct all experiments on NVIDIA V100 GPUs. Unless otherwise stated, we use 8$\times$8 GPUs for video experiments and 1$\times$8 GPUs for image experiments. Our default configurations follow the \emph{linear probing} settings in~\cite{chen-iccv21-mocov3,he-2021-mae}, which do \emph{not} utilize many common regularization strategies, such as mixup~\cite{zhang-2017-mixup}, cutmix~\cite{yun-iccv19-cutmix}, color jittering and so on. More details can be found in Appendix.

\subsection{Main Properties and Analysis}\label{sec:main_results}

\begin{table}[t]
\addtolength{\tabcolsep}{-4.5pt}
\renewcommand{\arraystretch}{1.2}
    \centering
    \caption{\textbf{Fine-tuning with self-supervised pre-trained model.} For tunable parameters, we also report the parameter percentage in the brackets. Besides, we report the top-1 accuracy on different dataset with the absolute value and the gap value relative to the \emph{full-tuning} regime. $^\dagger$ denotes $0.1\times$ learning rate due to unstable training.}
    \vspace{1.0pt}
    \begin{tabular}{l | c  |c c c | c c}
    \Xhline{1.0pt}
    \multirow{2}{*}{Method} & Avg. & \multicolumn{3}{c|}{\textbf{\textcolor{Image}{Image}}} & \multicolumn{2}{c}{\textbf{\textcolor{Video}{Video}}} \\
         &   Params (M)     & \textcolor{Image}{CIFAR-100} & \textcolor{Image}{SVHN} & \textcolor{Image}{Food-101} & \textcolor{Video}{SSv2} & \textcolor{Video}{HMDB51} \\
    \hline
    \baseline{Full-tuning}   & \baseline{86.04}~(100\%) & \baseline{85.90} & \baseline{97.67}$^\dagger$ & \baseline{90.09}$^\dagger$ & \baseline{53.97} & \baseline{46.41} \\
    Linear                   & 0.07~(0.08\%)  & 69.83~\bbb{-16.07} & 66.91~\bbb{-30.76} & 69.74~\bbb{-20.35} & 29.23~\bbb{-24.74} & 49.84~\aaa{+3.43} \\
    VPT~\cite{jia-2022-vpt}  & 0.08~(0.09\%)  & 82.44~\bbb{-3.46} & 94.02~\bbb{-3.65} & 82.98~\bbb{-7.11} & 43.73~\bbb{-10.24} & 52.67~\aaa{+6.26}
 \\
    \hline \hline
    \small{\ourabbr-1}                  & 0.10~(0.12\%) & 83.52~\bbb{-2.38} & 93.04~\bbb{-4.63} & 83.64~\bbb{-6.45} & 50.03~\bbb{-3.94} & 51.68~\aaa{+5.27} \\
    \small{\ourabbr-4}                  & 0.15~(0.17\%) & 84.83~\bbb{-1.07} & 96.19~\bbb{-1.48} & 85.42~\bbb{-4.67} & 54.70~\aaa{+0.73}& 51.81~\aaa{+5.40} \\
    \small{\ourabbr-64}                 & 1.26~(1.46\%) & 85.90~\aaa{0.00}  & 96.89~\bbb{-0.78} & 87.61~\bbb{-2.48} & 59.02~\aaa{+5.05} & 55.69~\aaa{+9.28}  \\
    \Xhline{1.0pt} 
    \end{tabular}
    \label{tab:ssl_pretrain}
\vspace{-1.5em}
\end{table}

We compare the performance of different fine-tuning approaches in Table~\ref{tab:ssl_pretrain} with the backbones pre-trained via the self-supervised paradigms. The results show that \ourabbr consistently surpasses linear probing and Visual Prompt tuning~(VPT) methods. Specifically, \ourabbr-64 outperforms VPT on image benchmark \textcolor{Image}{CIFAR-100}, \textcolor{Image}{SVHN}, and \textcolor{Image}{Food-101}, by 3.46\%, 2.87\%, and 4.63\% respectively. On the more challenging video action recognition dataset \textcolor{Video}{Something-Something~V2}, the superiority becomes even more significant, \ie, about 15\%. Note that even compared with the full fine-tuning strategy, our \ourabbr still outperforms by about 5\% Top-1 accuracy on \textcolor{Video}{SSv2} dataset.
To summarize, our \ourabbr is highly parameter-efficient, as well as yielding good performance with parameter size at most 2\% times than the full fine-tuning manner.

\subsection{Scaling Tunable Parameters Up} 
\begin{figure}[t]
\begin{center}
    \begin{minipage}{0.5\textwidth}
    \includegraphics[width=\textwidth]{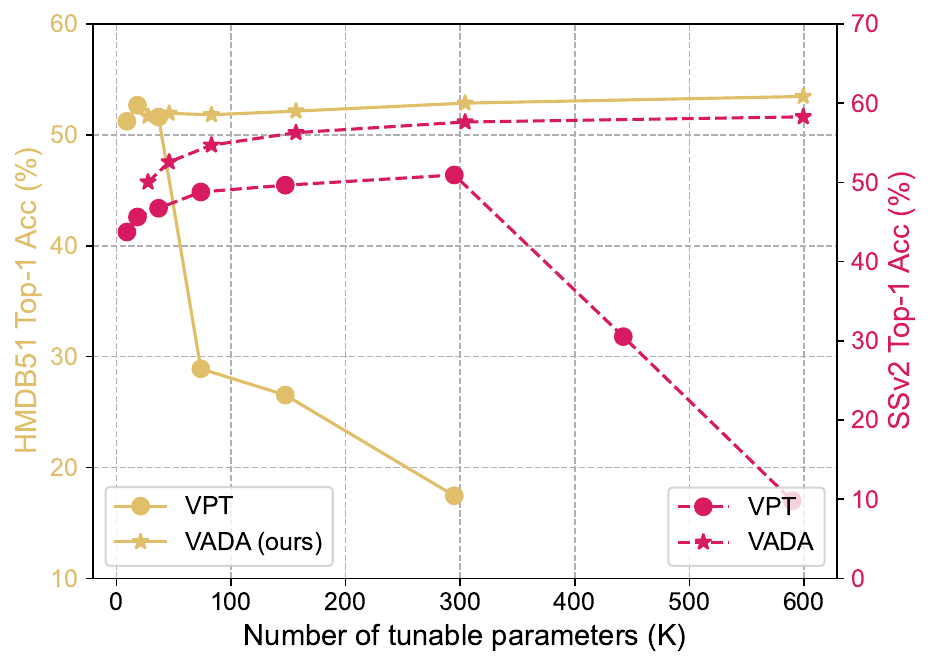}
    \caption{\textbf{The trend of performance as the number of tunable parameters grows up.}  The accuracy of VPT drops dramatically when the parameter number exceeds task-specific value, while \ourabbr is robust to the increasing parameters.}\label{fig:params_acc}
    \end{minipage}
    \hspace{0.02\textwidth}
    \begin{minipage}{0.46\textwidth}
    \includegraphics[width=\textwidth]{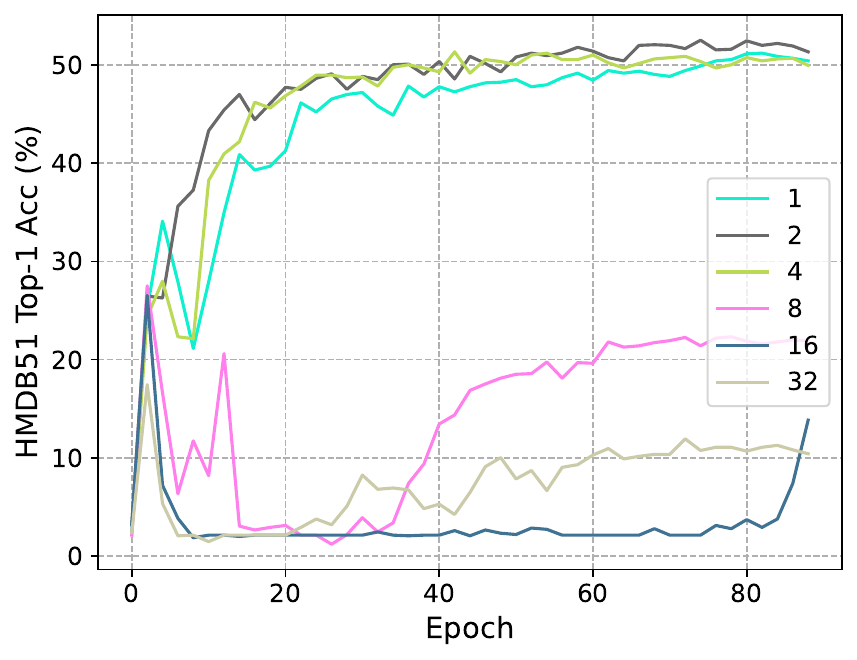}
    \caption{\textbf{Test accuracy of VPT~\cite{jia-2022-vpt} with different number of introduced tokens.} The optimization procedure becomes unstable when the token number is equal or larger than eight on HMDB51 dataset~\cite{kuehne-iccv11-hmdb}.}
    \label{fig:epoch_top1}
    \end{minipage}
\end{center}
\vspace{-2.5em}
\end{figure}

Even though there are only limited parameters introduced, one might also argue that more tunable parameters of \ourabbr contribute to its higher accuracy compared with VPT~\cite{jia-2022-vpt}. We conduct experiments to make a comprehensive discussion on this aspect.

As described in Sec.~\ref{sec:num_parameters}, the number of tunable parameters can be adjusted by changing the number of introduced tokens for VPT, or the hidden feature dimension for \ourabbr. As shown in Figure~\ref{fig:params_acc}, we conduct experiments with a wide range of tunable parameters on both \textcolor{Video}{SSv2} and \textcolor{Video}{HMDB-51} datasets. Since \ourabbr and VPT share the same number of parameters of classification head on a specific dataset, we only report the tunable parameters on the x-axis, which comes from the visual prompts~(VPT) or weight/bias of the down-up fully-connected layers~(\ourabbr), without calculating the parameters of classification head. For VPT, the number of introduced tokens is chosen from \{1, 2, 4, 8, 16, 32, 48, 64\}. Similarly, the number of hidden dimensions in AdaptFormer is in \{1, 2, 4, 8, 16, 32\}. \ourabbr has a slight performance gain or maintains the accuracy stably when the parameters scale up. On the contrary, the performance of VPT decreases dramatically when the parameters exceed the task-specific value. Moreover, choosing the most suitable number of token number becomes laborious since it might be task-specific (\ie varying from one dataset to the other one). For example, the accuracy of VPT keeps going up when the number of tunable parameters increases up to 300K on SSv2, whereas it begins to drop when the number of tunable parameters exceeds 50K on HMDB-51.

We further study the optimization procedures of VPT by monitoring the test accuracy of the training stage. As shown in Figure~\ref{fig:epoch_top1}, we gradually increase the number of tokens in VPT and plot the Top-1 accuracy of each epoch. The training stages are stable when the number of tokens is less than or equal to 4, \eg, \{1, 2, 4\}. However, when the number becomes 8 or larger, \eg, \{8, 16, 32\}, the training procedure collapses at about the tenth epoch and achieves poor performance at the end of the training stage. On the contrary, the optimization procedures of \ourabbr are stable when the number of parameters varies across a large range, as shown in Table~\ref{tab:hidden_dim}. The top-1 accuracy fluctuates within 1.5\% when the number of parameters increases from 0.44M~(\texttt{dim=16}) to 4.87M~(\texttt{dim=256}).

\subsection{Multi-Label Classification}

We further conduct experiments on dataset with larger scale and diversity. Specifically, we evaluate \ourabbr on NUS-WIDE~\cite{chua2009nus} for multi-label classification. 
NUS-WIDE contains 269,648 images collected from Flicker, which are annotated with 81 visual concepts. Since some images are not available on Flicker, we only use 220,000 images following~\cite{ben2020asymmetric, durand2019learning}. 
We utilize mean average precision~(mAP) as performance metric.

\noindent\textbf{Settings and results.} Our training settings mainly follow ASL~\cite{ben2020asymmetric}. Specifically, We trained all models for 40 epochs using Adam optimize and 1-cycle learning rate policy~\cite{smith2018disciplined}. The maximal learning rate is 0.001.
As shown in Table~\ref{tab:multi_label}, though AdaptFormer-64 achieves a slightly lower mAP than fine-tuning, it significantly reduces the amount parameters that need to be updated~(from 85.86 to 1.25M). Moreover, AdaptFormer has an clear advantage over other fine-tuning approaches including linear probing and VPT.

\begin{table}[t]
\addtolength{\tabcolsep}{-4.5pt}
\renewcommand{\arraystretch}{1.2}
    \centering
    \caption{\rebuttal{\textbf{AdaptFormer for multi-label classification.} }}
    \vspace{1.0pt}
    \begin{tabular}{l | c  |c}
    \Xhline{1.0pt}
    Method & Params (M) & NUS-WIDE~\cite{chua2009nus} \\
    \hline
    \baseline{Full-tuning}   & \baseline{85.86}~(100\%) & \baseline{61.26} \\
    Linear                   & 0.06~(0.08\%)  & 51.19~\bbb{-27.25}   \\
    VPT~\cite{jia-2022-vpt}  & 0.07~(0.09\%)  & 57.08~\bbb{-7.56}   \\
    \hline \hline
    \small{\ourabbr-1}                  & 0.09~(0.12\%) & 57.51~\bbb{-4.08}   \\
    \small{\ourabbr-4}                  & 0.15~(0.17\%) & 58.14~\bbb{-2.13}   \\
    \small{\ourabbr-64}                 & 1.25~(1.46\%) & 59.07~\bbb{-0.06}   \\
    \Xhline{1.0pt} 
    \end{tabular}
    \label{tab:multi_label}
\end{table}

\subsection{Ablation Studies}\label{sec:abla}

We ablate our \ourabbr to study what properties make for a good \ourabbr and observe several intriguing properties. The ablation studies conducted in this work are all performed on the SSv2 validation set~\cite{goyal-iccv17-sthsthv2}.

\begin{table}[ht]
\vspace{-1.0em}
\centering
\caption{\textbf{\ourabbr ablation experiments} with ViT-B/16 on \textcolor{Video}{SSv2}. We report the top-1 accuracy on the \texttt{val} set. Most suitable settings are marked in \colorbox{bestcolor}{color}.}\label{tab:ablations}
\subfloat[
\textbf{Middle dimension $\hat{d}$.}\label{tab:hidden_dim}
]{
\begin{minipage}{0.3\textwidth}
{\begin{center}
\tablestyle{2pt}{1.13}
\begin{tabular}{x{40}x{30}x{30}}
\Xhline{1.0pt}
mid dim &  \#params& top-1 \\
\hline
1   & 0.16M & 50.03 \\
16  & 0.44M & 57.62 \\
32  & 0.73M & 58.27 \\
\bestcell{64}  & \bestcell{1.32M} & \bestcell{\textbf{59.02}} \\
256 & 4.87M & 58.87 \\
\Xhline{1.0pt}
\end{tabular}
\end{center}}\end{minipage}
}
\subfloat[
\textbf{AdaptMLP inserted layers and form.}\label{tab:insert_form}
]{
\begin{minipage}{0.4\textwidth}{\begin{center}
\tablestyle{0.1pt}{1.33}
\begin{tabular}{y{40}x{40}x{30}x{30}}
\Xhline{1.0pt}
layers & form & \#params & top-1 \\
\hline
1 $\rightarrow$ 6 & parallel & 0.73 & 50.48 \\
7 $\rightarrow$ 12 & parallel  & 0.73 & 57.99 \\
\bestcell{1 $\rightarrow$ 12} & \bestcell{parallel} & \bestcell{1.32} & \bestcell{59.02} \\
1 $\rightarrow$ 12 & sequential & 1.32 & 58.17 \\
\Xhline{1.0pt}
\end{tabular}

\end{center}}
\vspace{-1.5em}
\end{minipage}
}
\subfloat[
\textbf{Scaling factor $s$.}\label{tab:s_factor}
]{
\begin{minipage}{0.2\textwidth}{\begin{center}
\tablestyle{3pt}{1.33}
\begin{tabular}{x{20}x{30}}
\Xhline{1.0pt}
factor & top-1 \\
\hline
0.01 & 53.44 \\
0.05 & 58.85 \\
\bestcell{0.10} & \bestcell{59.02} \\
0.20 & 58.89 \\
\Xhline{1.0pt}
\end{tabular}
\end{center}}\end{minipage}
}
\end{table}

\noindent\textbf{Middle dimension.}
The middle dimension controls the number of introduced parameters by \ourabbr. Lower middle dimensions introduce fewer parameters with a possible performance cost.  We ablate \ourabbr on the middle feature dimension to study this effects. As shown in Table~\ref{tab:hidden_dim}, the accuracy consistently improves when the middle dimension increases up to 64 and reaches the saturation point when the middle dimension is about 64 on SSv2 dataset.
We note that our \ourabbr can achieve a decent performance when the middle dimension reduces even to one, about 50.03\% top-1 accuracy.

We conduct more extensive ablation studies on middle dimension in Appendix Table~\ref{tab:mid_dimension} and found that the optimal middle dimension varies per dataset. For example, the accuracy reaches saturation when the middle dimension equals 64 on SSv2, whereas for NUS-WIDE dataset, the mAP slightly improves when the middle dimension increases from 64 to 512. However, AdaptFormer with middle dimension as 512 has 0.75 mAP higher (59.82 vs. 59.07 mAP) than the one with 64 at the cost of about 8 times more parameters. Therefore, we choose the \texttt{middle dimension=64} for both SSv2 and NUS-WIDE for a better trade-off.

\rebuttal{
\noindent\textbf{Scaling factor.} The scaling factor $s$ is introduced to balance the \textit{task-agnostic} features~(generated by the original frozen branch) and the \textit{task-specific} features~(generated by the tunable bottleneck branch). We evaluate \ourabbr with multiple $s$ values and the results are summarized in Table~\ref{tab:s_factor}. Different from the scaling factor in NLP field which prefer $s$ larger than 1~(\eg, $s=4$ in~\cite{he2022towards}), we empirically found that the $s$ should be $<1$ for vision tasks, otherwise the fine-tuning would become unstable. Besides, we found that \ourabbr achieves optimal performance with $s=0.1$. A larger or smaller $s$ would bring slight performance drop. Thus, we choose $s=0.10$ as a default setting. 
}

\noindent\textbf{\ourabbr position.} As shown in Table~\ref{tab:insert_form}, we further ablate on the specific position to introduce the AdaptMLP block. We gradually increase the number of AdaptMLP layers with a step of three (\texttt{start}~$\rightarrow$~\texttt{end}, both included). We observe that the performance of \ourabbr has a positive correlation with the number of added layers. In addition, \ourabbr prefers the 
top part (the one far away from the input image) of the network to the bottom part when introducing the same number of layers, \eg, \ourabbr with \texttt{7}~$\rightarrow$~\texttt{12} obtains over 14.5\% higher accuracy than \texttt{1}~$\rightarrow$~\texttt{6}, though both equipped with six AdaptMLP layers.

\begin{figure}[ht]
    \centering
    \begin{minipage}{0.4\textwidth}
    \vspace{-10pt}
    \includegraphics[width=\textwidth]{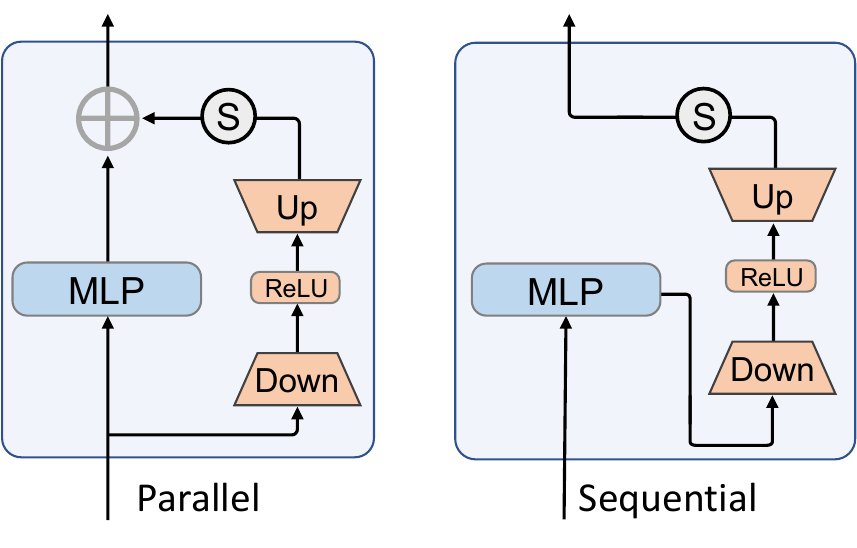}
    \caption{\textbf{Illustration of the \emph{parallel} and \emph{sequential} insertion form}. Comparison results are shown in Table~\ref{tab:insert_form}.}\label{fig:insertion_form}
    \label{fig:abl_parallel_sequential}
    \end{minipage}
    \hfill
    \begin{minipage}{0.4\textwidth}
    \includegraphics[width=\textwidth]{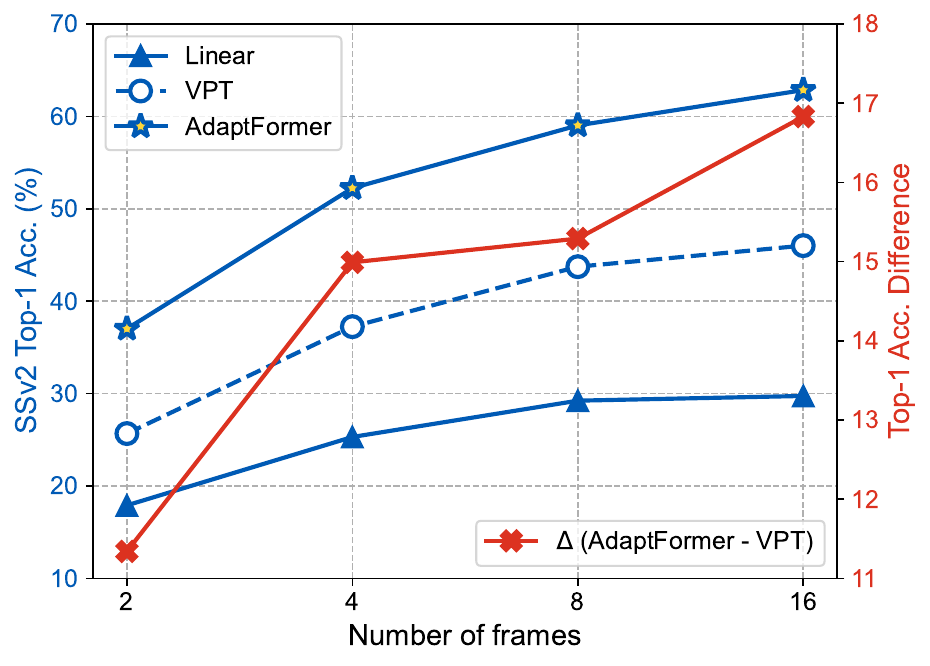}
    \caption{\textbf{Performance with video frames number.} AdaptFormer outperforms VPT and linear fine-tuning.}
    \label{fig:frame_acc}
    \end{minipage}
\vspace{-1.5em}
\end{figure}

\noindent\textbf{Insertion form.} We study the insertion formulation by comparing the \emph{parallel} and \emph{sequential} instances which are illustrated in Figure~\ref{fig:insertion_form}. As shown in Table~\ref{tab:insert_form}, the parallel \ourabbr is able to outperform the sequential one by 0.85\% top-1 accuracy. The reason might be: \textbf{(1)} the parallel design maintains the original feature using an independent branch and aggregating updated context by element-wise scaled sum; \textbf{(2)} the sequential design is equivalent to adding more layers, which might cause optimization difficulty. Therefore, we adopt the parallel design as our default setting due to its superiority. 

\vspace{-1.5em}

\noindent\textbf{Number of frames.} The number of embedded patch tokens increases linearly with the number of video frames for the plain ViT~\cite{dosovitskiy-2020-vit}. We conduct experiments with the different number of frames, \ie, \{2, 4, 8\} and the results are shown in Figure~\ref{fig:frame_acc}. We observe that increasing the number of frames is beneficial for all these three fine-tuning methods. However, AdaptFormer consistently outperforms the linear manner (\eg, +30\% top-1 accuracy on 8 input frames) and VPT method(\eg, +14\% top-1 accuracy on 8 input frames).

\vspace{-0.8em}
\subsection{Towards Visual Recognition Generalist Agent}
\vspace{-0.8em}

In the above experiments, we typically utilize a modality-specific pre-trained checkpoint for the corresponding downstream tasks. For example, we use \textcolor{Video}{Kinetics-400}~(\textcolor{Video}{\textbf{video domain}}) pre-trained model for downstream video action recognition on \textcolor{Video}{Something-Something~V2} and \textcolor{Video}{HMDB-51} benchmarks. Besides, we use \textcolor{Image}{ImageNet-21K}~(\textcolor{Image}{\textbf{image domain}}) pre-rained model for downstream image classification on \textcolor{Image}{CIFAR-100}, \textcolor{Image}{SVHN} and \textcolor{Image}{Food-101} benchmarks. Our \ourabbr achieves superior performances in this \emph{same network with modality-specific weights} scenario.
\begin{wraptable}{r}{7cm}
\vspace{-1.0em}
\addtolength{\tabcolsep}{1.0pt}
\renewcommand{\arraystretch}{1.1}
    \centering
    \caption{\textbf{Fine-tuning on \textcolor{Video}{\textbf{video}} data with \textcolor{Image}{\textbf{image}}} pre-trained model.}\label{tab:image2video}
    \begin{tabular}{l | c  | c}
    \Xhline{1.0pt}
    \multirow{2}{*}{Method} & Avg.  & Fine-tuning \\
         &   Params (M) &   \textcolor{Video}{SSv2} \\
    \hline
    Full-tuning              & 86.36   & 41.50 \\
    Linear                   & 0.15  & 6.56    \\
    VPT~\cite{jia-2022-vpt}  & 0.16  & 16.94 \\
    \hline
    \ourabbr                 & 1.33  & 46.06 \\
    \Xhline{1.0pt} 
    \end{tabular}
\end{wraptable}
Next, we take a further step to ask what would happen if using \emph{the same network with the modality-agnostic weights} for multiple tasks in the multi-modalities downstream tasks?

We use the model pre-trained on \textcolor{Image}{ImagNet-21k} to do action recognition on \textcolor{Video}{SSv2}. As shown in Table~\ref{tab:image2video}, \ourabbr is robust to domain shift caused by modality. The experimental results show that the linear probe approach obtains a very poor accuracy (\ie, 6.56\% top-1 accuracy) when fine-tuning on \textcolor{Video}{SSv2}. Meanwhile, VPT~\cite{jia-2022-vpt} achieves a better performance than linear probe but it is not decent~(\ie, 16.94\% top-1 accuracy). Our \ourabbr, compared to the above two methods, attains a promising 46.06\% top-1 accuracy, which is even higher than the full-tuning schedule (+4.56\%). 
\vspace{-0.8em}
\subsection{Visualization}

\begin{figure}[ht]
\vspace{-1.0em}
    \centering
    \begin{subfigure}{0.24\textwidth}
    \includegraphics[width=\textwidth]{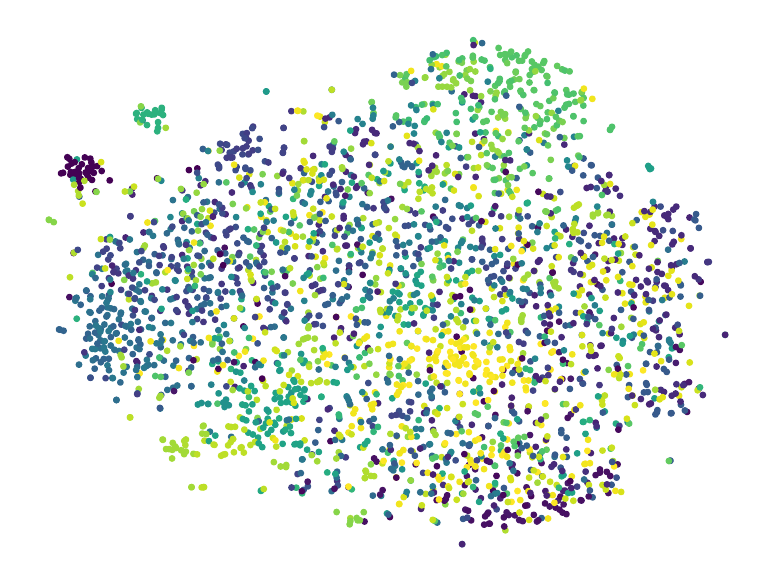}
    \caption{Linear (\textcolor{blue}{0.08\%})\\\hspace*{4mm}(\textcolor{red}{Top1 29.23\%})}
    \end{subfigure}
    \begin{subfigure}{0.24\textwidth}
    \includegraphics[width=\textwidth]{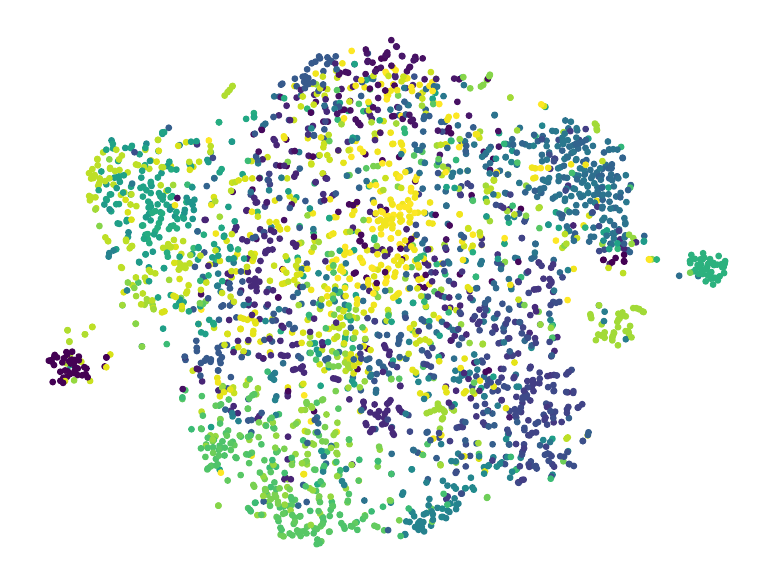}
    \caption{VPT (\textcolor{blue}{0.09\%})\\\hspace*{3mm}(\textcolor{red}{Top1 43.73\%})}
    \end{subfigure}
    \begin{subfigure}{0.24\textwidth}
    \includegraphics[width=\textwidth]{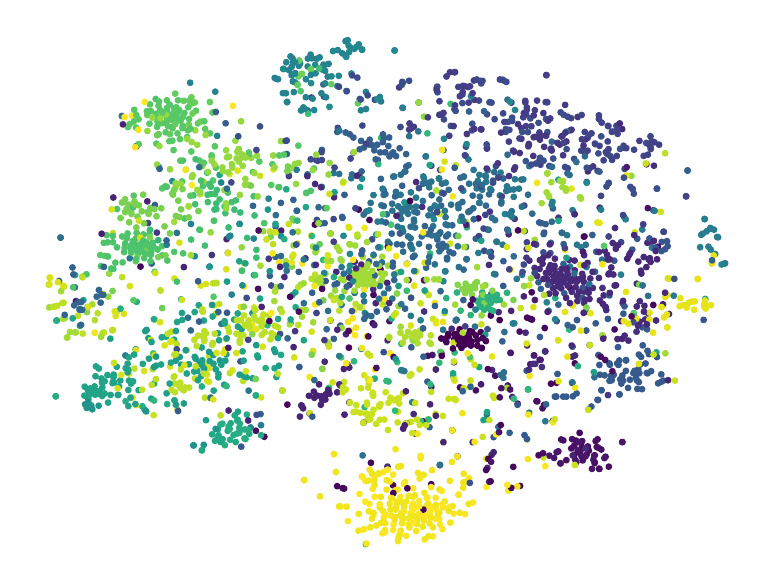} 
    \caption{Full fine-tune (\textcolor{blue}{100\%})\\\hspace*{8mm}(\textcolor{red}{Top1 53.97\%})}
    \end{subfigure}
    \begin{subfigure}{0.24\textwidth}
    \includegraphics[width=\textwidth]{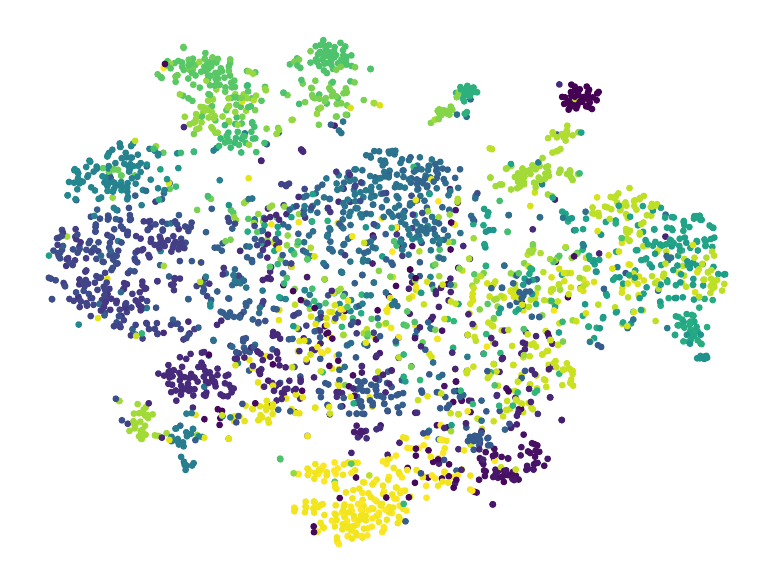}
    \caption{AdaptFormer (\textcolor{blue}{1.26\%})\\\hspace*{8mm}(\textcolor{red}{Top1 59.02\%})}
    \end{subfigure}
\caption{\textbf{t-SNE visualizations on \textcolor{Video}{SSv2} val dataset.} We extract the final classification features from the top linear layer for t-SNE visualizations. \rebuttal{The \textcolor{red}{top-1 accuracy} is reported in red, while the \textcolor{blue}{relative parameter} (compared to the full fine-tuning strategy) is reported in blue.}}\label{fig:tsne-video}
\vspace{-1.0em}
\end{figure}

To evaluate the quality of the produced features, we conduct t-SNE~\cite{van2008visualizing} visualizations on AdaptFormer and other baseline methods. The features are extracted from the SSv2 validation set via the ViT-Base backbone. Figure~\ref{fig:tsne-video} shows that the linear fine-tuning and the VPT methods tend to output mixed features as shown in Figure~\ref{fig:tsne-video}(a)-(b). Compared with the above two methods, the full fine-tuning strategy performs well in projecting features. However, it consumes huge computational sources to tune the whole network parameters.
Figure~\ref{fig:tsne-video}(d) validates that our AdaptFormer facilitates ViT-Base in generating more separable representations with fewer learnable parameters.

\vspace{-0.5em}
\section{Conclusion}
\vspace{-0.5em}
We present a conceptually simple yet effective framework, \ourabbr, for efficiently adapting a pre-trained Vision Transformer (ViT) backbone to scalable vision recognition tasks. By introducing AdaptMLP, our \ourabbr is able to fine-tune the lightweight modules for producing features adapted to multiple downstream tasks. The extensive experiments on five datasets, covering both the image and the video domains, validate that our proposed methods are able to increase the ViT’s transferability with little computational cost. We hope our work will inspire future research in exploring more efficient fine-tuning methods for large vision models. One limitation is that \ourabbr is only employed in recognition tasks in this work, it's unclear whether it can work well in tasks beyond recognition, \eg, object detection and semantic segmentation. We leave it for the future exploration.
Since our method is specially designed for efficient fine-tuning, we do not foresee obvious undesirable ethical/social impacts at this moment.

\noindent\textbf{Acknowledgment.} This work is supported by CCF-Tencent Open Fund. Ping Luo is supported by the General Research Fund of HK No.27208720, No.17212120, and No.17200622.


{\small
\bibliographystyle{plain}
\bibliography{ref}
}
\clearpage
\section*{Checklist}

\begin{enumerate}

\item For all authors...
\begin{enumerate}
  \item Do the main claims made in the abstract and introduction accurately reflect the paper's contributions and scope?
    \answerYes{}
  \item Did you describe the limitations of your work?
   \answerYes{\rebuttal{Shown in \textbf{Conclusion} Section}.}
  \item Did you discuss any potential negative societal impacts of your work?
  \answerYes{\rebuttal{Shown in \textbf{Conclusion} Section}.}
  \item Have you read the ethics review guidelines and ensured that your paper conforms to them?
   \answerYes{}
\end{enumerate}

\item If you are including theoretical results...
\begin{enumerate}
  \item Did you state the full set of assumptions of all theoretical results?
    \answerNA{}
        \item Did you include complete proofs of all theoretical results?
    \answerNA{}
\end{enumerate}

\item If you ran experiments...
\begin{enumerate}
  \item Did you include the code, data, and instructions needed to reproduce the main experimental results (either in the supplemental material or as a URL)?
   \answerYes{As a URL shown in the abstract.}
  \item Did you specify all the training details (e.g., data splits, hyperparameters, how they were chosen)?
    \answerYes{Shown in supplementary materials.}
        \item Did you report error bars (e.g., with respect to the random seed after running experiments multiple times)?
    \answerNA{}
        \item Did you include the total amount of compute and the type of resources used (e.g., type of GPUs, internal cluster, or cloud provider)?
     \answerYes{Please see Section~\ref{Sec:exp_setting}}
\end{enumerate}

\item If you are using existing assets (e.g., code, data, models) or curating/releasing new assets...
\begin{enumerate}
  \item If your work uses existing assets, did you cite the creators?
    \answerYes{}
  \item Did you mention the license of the assets?
   \answerYes{Shown in supplementary materials.}
  \item Did you include any new assets either in the supplemental material or as a URL?
   \answerNo{}
  \item Did you discuss whether and how consent was obtained from people whose data you're using/curating?
  \answerYes{We used publicly available datasets whose licenses allow research usage.}
  \item Did you discuss whether the data you are using/curating contains personally identifiable information or offensive content?
   \answerNo{To the best of our knowledge, the data we used contains no personally identiﬁable information or offensive content.}
\end{enumerate}

\item If you used crowdsourcing or conducted research with human subjects...
\begin{enumerate}
  \item Did you include the full text of instructions given to participants and screenshots, if applicable?
  \answerNA{}
  \item Did you describe any potential participant risks, with links to Institutional Review Board (IRB) approvals, if applicable?
  \answerNA{}
  \item Did you include the estimated hourly wage paid to participants and the total amount spent on participant compensation?
  \answerNA{}
\end{enumerate}

\end{enumerate}
\clearpage

\appendix
\section{Appendix}
In this supplementary material, we will include the details about the pre-training and fine-tuning processes, the extensive experiments of \ourabbr on hierarchical vision transformers~(\eg, \ourabbr-Swin), and the pseudo-code of AdaptMLP in a PyTorch-like style.

\subsection{Experimental Settings}

\subsubsection{Pre-training Approaches}\label{Sec:joint-setting}

\noindent\textcolor{Image}{\textbf{Image.}} We use MAE~\cite{he-2021-mae} as our self-supervised pre-training method in the \textcolor{Image}{image} domain, a simple yet effective method that first masks nearly 75\% patches of the input image and then reconstructs the missing pixels. Specifically, we directly adopt the checkpoint\footnotemark of ViT-B/16 for convenience,
\footnotetext{\small \url{https://dl.fbaipublicfiles.com/mae/pretrain/mae_pretrain_vit_base.pth}} which is pre-trained on ImageNet-1K~\cite{deng-cvpr09-imagenet} for 800 epochs.

\noindent\textbf{\textcolor{Video}{Video.}} We use VideoMAE~\cite{tong-2022-videomae} as our self-supervised pre-training method in the \textcolor{Video}{video} domain, which is an direct extension of MAE to the video domain. VideoMAE utilizes the plain ViT~\cite{dosovitskiy-2020-vit} architecture of joint space-time attention mechanism~\cite{arnab2021vivit, liu2021video} and an extremely high proportion of masking ratio~(\ie, 90\% to 95\%) for pre-training. We also directly use the publicly available checkpoint\footnotemark\footnotetext{\small \url{https://drive.google.com/file/d/1tEhLyskjb755TJ65ptsrafUG2llSwQE1/view?usp=sharing}}, which is pre-trained on Kinetics-400~\cite{carreira2017quo}.

\subsubsection{Implementation Details of Fine-tuning}

\begin{table}[ht]
\renewcommand{\arraystretch}{1.2}
    \centering
    \caption{\textbf{Fine-tuning settings.} We present the shared configurations, like the optimizer and the base learning rate, the upper part, and show the seperated ones in the lower part. }\label{tab:settings}
    \vspace{1pt}
\begin{tabular}{l  c  c}
\Xhline{1.0pt}
Configuration & \textcolor{Image}{Image}  & \textcolor{Video}{Video}\\
\hline
optimizer & \multicolumn{2}{c}{SGD}  \\
base learning rate & \multicolumn{2}{c}{0.1} \\
weight decay & \multicolumn{2}{c}{0} \\
optimizer momentum & \multicolumn{2}{c}{0.9} \\
batch size & \multicolumn{2}{c}{1024 images/frames}  \\
learning rate schedule & \multicolumn{2}{c}{cosine decay~\cite{loshchilov2016sgdr}} \\
GPU numbers     & 8 & 64 \\
warmup epochs & 20  & 10 \\
training epochs & 100 & 90 \\
augmentation & RandomResizedCrop~\cite{he-2021-mae}  & MultiScaleCrop~\cite{tong-2022-videomae} \\
\Xhline{1.0pt}
\end{tabular}

\end{table}

The implementation details are summarized in Table~\ref{tab:settings}. 
The \textcolor{Video}{video} experiments are conducted on 64 Tesla V100 GPUs, while the \textcolor{Image}{image} experiments are performed on 8 Tesla V100 GPUs. For the optimizer, different from~\cite{you2017large} that adopts LARS, we leverage SGD~\cite{sutskever2013importance} for stable training on small-scale dataset~(\eg, \textcolor{Image}{CIFAR10}). The actual learning rate is calculated by:
\textit{lr} = \textit{base\_lr}$\times$batchsize / 256 following the linear \textit{lr} scaling rule~\cite{goyal-2017-accurate}. More detailed training configurations are presented in Table~\ref{tab:settings}, including the batchsize, learning rate schedule and etc.

The experimental settings of \textcolor{Image}{image} and \textcolor{Video}{video} mainly follow the ones utilized in MAE~\cite{he-2021-mae} and VideoMAE~\cite{tong-2022-videomae}, respectively. We insert an extra BatchNorm layer~\cite{ioffe2015batch} without affine transformation~(\ie  ~\texttt{affine=False}) before the final fully connected layer, following the common practice to normalize the pre-trained features~\cite{doersch2015unsupervised, he-2021-mae}. In addition, there is \emph{no} flip augmentation during the fine-tuning stage for video data.

\subsection{More Supplementary Results}

\subsubsection{AdaptFormer with Supervised Pre-training}

In addition to the self-supervised pre-training presented in the main paper, we also evaluate \ourabbr with the supervised pre-trained model. The results in Table~\ref{tab:super_pretrain} show that \ourabbr still outperforms linear probe and VPT obviously. In addition, \ourabbr surpasses full-tuning on four benchmarks~(\textcolor{Image}{CIFAR100}, \textcolor{Image}{SVHN}, \textcolor{Video}{SSv2}, \textcolor{Video}{HMDB51}) with only 1.46\% parameters. On the remaining  benchmark~(\textcolor{Image}{Food-101}), \ourabbr achieves an almost comparable performance to full-tuning~(90.89\% \emph{v.s.} 90.96\%).

\begin{table}[h]

\addtolength{\tabcolsep}{-4.5pt}
\renewcommand{\arraystretch}{1.2}
    \centering
    \caption{\textbf{Fine-tuning with \emph{supervised} pre-trained model.} We report the tunable parameters percentage in the brackets. Besides, we report the top-1 accuracy on different dataset with the absolute value and the gap value relative to the \emph{full-tuning} regime.
    }
    \vspace{1.0pt}
    \begin{tabular}{l | c  |c c c | c c}
    \Xhline{1.0pt}
    \multirow{2}{*}{Method} & Avg. & \multicolumn{3}{c|}{\textbf{\textcolor{Image}{Image}}} & \multicolumn{2}{c}{\textbf{\textcolor{Video}{Video}}} \\
         &   Params (M)     & \textcolor{Image}{CIFAR-100} & \textcolor{Image}{SVHN} & \textcolor{Image}{Food-101} & \textcolor{Video}{SSv2} & \textcolor{Video}{HMDB51} \\
    \hline
    \baseline{Full-tuning}   & \baseline{86.04}~(100\%) & \baseline{89.12} & \baseline{95.41} & \baseline{90.96} & \baseline{53.62} & \baseline{59.38} \\
    Linear                   & 0.07~(0.08\%)  & 85.95~\bbb{-3.17} & 55.36~\bbb{-40.05} & 88.14~\bbb{-2.82} & 35.49~\bbb{-18.13} & 70.31~\aaa{+10.93} \\
    VPT~\cite{jia-2022-vpt}  & 0.08~(0.09\%)  & 90.97~\aaa{+1.85} & 92.77~\bbb{-2.64} & 90.16~\bbb{-0.80} & 55.22~\aaa{+1.60} & 71.56~\aaa{+12.18}
 \\
    \hline \hline
    \small{\ourabbr-64}                 & 1.26~(1.46\%) & 91.86~\aaa{+2.73} & 97.29~\aaa{+1.88} & 90.89~\bbb{-0.07} & 60.18~\aaa{+6.56} & 73.21~\aaa{+13.83}  \\
    \Xhline{1.0pt} 
    \end{tabular}
    \label{tab:super_pretrain}

\end{table}

\subsubsection{AdaptFormer on Swin Transformer}

\noindent \textbf{Settings.} We further demonstrate the effectiveness of \ourabbr on hierarchical vision transformers, \eg, Swin~\cite{liu-iccv21-swin, liu2021video}. We name \ourabbr applied to Swin as \emph{\ourabbr-Swin}, to distinguish plain \ourabbr (without any suffix) which is applied to the vanilla ViT~\cite{dosovitskiy-2020-vit}. It is noted that we can adopt AdaptMLP to Swin easily without any special modification as Swin and ViT share the same MLP architecture. However, VPT~\cite{jia-2022-vpt} needs additional handcraft designs to be suitable for the shifted local windows in the prevalent hierarchical vision transformers, which hinders its general applications.

We utilize Swin-B~\cite{liu-iccv21-swin} and the video counterpart~\cite{liu2021video} for \textcolor{Image}{image} and \textcolor{Video}{video}. Similarly, we also directly use the officially provided checkpoints\footnotemark, \footnotetext{\noindent\small Image: \url{https://github.com/SwinTransformer/storage/releases/download/v1.0.4/swin_base_patch244_window877_kinetics600_22k.pth} \\
\text{~~~} Video: \url{https://github.com/SwinTransformer/storage/releases/download/v1.0.0/swin_base_patch4_window7_224_22k.pth}}
which are pre-trained on ImageNet-21K~\cite{deng-cvpr09-imagenet} and Kinetics-600~\cite{carreira2018short}.

\begin{table}[ht]
\addtolength{\tabcolsep}{-5.2pt}
\renewcommand{\arraystretch}{1.2}
    \centering
    \caption{\textbf{Fine-tuning with \emph{Swin} Transformer}. We utilize Swin-B~\cite{liu-iccv21-swin} and Video Swin-B~\cite{liu2021video} for \textcolor{Image}{\textbf{image}} and \textcolor{Video}{\textbf{video}} experiments, respectively. Parameter percentage and  performance difference are reported relative to \emph{full-tuning} schedule. 
    }
    \vspace{1.0pt}
    \begin{tabular}{l | c  |c c c | c c}
    \Xhline{1.0pt}
    \multirow{2}{*}{Method} & Avg. & \multicolumn{3}{c|}{\textbf{\textcolor{Image}{Image}}} & \multicolumn{2}{c}{\textbf{\textcolor{Video}{Video}}} \\
         &   Params (M)     & \textcolor{Image}{CIFAR-100} & \textcolor{Image}{SVHN} & \textcolor{Image}{Food-101} & \textcolor{Video}{SSv2} & \textcolor{Video}{HMDB51} \\
    \hline
    \baseline{Full-tuning}   & \baseline{87.19}~(100\%) & \baseline{89.95} & \baseline{97.03} & \baseline{91.43} & \baseline{52.92} & \baseline{68.73} \\
    Linear                   & 0.11~(0.13\%)  & 89.07~\bbb{-0.88} & 69.06~\bbb{-27.97} & 90.64~\bbb{-0.79} & 28.32~\bbb{-24.61} & 74.00~\aaa{+5.27} \\
    \hline \hline
    \small{\ourabbr-Swin}     & 1.25~(1.43\%) & 91.88~\aaa{+1.93} & 97.31~\aaa{+0.28} & 91.86~\aaa{+0.43} & 54.09~\aaa{+1.17} & 74.65~\aaa{+5.92}  \\
    \Xhline{1.0pt} 
    \end{tabular}
    \label{tab:swin}
\vspace{-1em}
\end{table}

\noindent\textbf{Results.} Since VPT is not applicable in Swin, we do not report its performance.  Table~\ref{tab:swin} shows \ourabbr-Swin performs well 
compared with other tuning strategies. For \textcolor{Image}{image} benchmarks, our method can outperform full-tuning approach with only 1.43\% parameters. Moreover, \ourabbr-Swin surpasses linear probing by a significant margin, especially on the challenging dataset, \textcolor{Video}{SSv2}. The results validate that \ourabbr is able to generally boost the transferability of various vision Transformer variants.

\subsection{\rebuttal{Possible Architectures}}
\rebuttal{
We explore other possible architectures utilized in AdaptFormer. Specifically, we further replace the MLP architectures within the AdaptMLP module by the convolution layer, depthwise convolution layer, and LayerNorm layer. For fair comparisons, we carefully design the above modules to meet the comparable number of parameters (\~ 1.3M). The experimental results of different adapter modules are shown in Table~\ref{tab:architectures}, which validates that the simple MLP modules are simple yet effective compared with the other architectures. For example, our AdaptMLP module surpasses the AdaptConv module by 0.55\% Top1 accuracy on SSv2 dataset.
}

\begin{table}[ht]
\addtolength{\tabcolsep}{-2.2pt}
\renewcommand{\arraystretch}{1.2}
    \centering
    \caption{\textbf{Fine-tuning with different adapter modules}. We use AdaptConv to denote the designed adapter module with convolution layers, while AdaptDepthwise-Conv is utilized to denote the designed adapter module with depthwise convolution layers. Besides, we also replace the MLP architectures with LayerNorm layer as AdaptLayerNorm-In.
    }
    \vspace{1.0pt}
    \begin{tabular}{l | c  |c c c }
    \Xhline{1.0pt}
    Methods & Avg Parameters & SSv2 Top1 & NUS-WIDE mAP & CIFAR100 Top1\\
    \hline
    \baseline{AdaptMLP}   & \baseline{1.28} & \baseline{59.02} & \baseline{59.07} & \baseline{85.93} \\
    \hline \hline
    AdaptConv                   & 1.39  & 58.47 & 58.86 & 85.42 \\
    AdaptDepthwise-Conv         & 1.29  & 58.15 & 58.73 & 85.37 \\
    AdaptLayerNorm-In           & 1.30  & 57.85 & 58.51 & 85.71 \\
    \Xhline{1.0pt} 
    \end{tabular}
    \label{tab:architectures}
\vspace{-1em}
\end{table}

\subsection{\rebuttal{Evaluation on ImageNet-1k datasets}}
\begin{table}[ht]
\addtolength{\tabcolsep}{-0.2pt}
\renewcommand{\arraystretch}{1.2}
    \centering
    \caption{\textbf{Fine-tuning with AdaotFormer on ImageNet-1k dataset}. We load the weights pretrained on ImageNet-21K and evaluate the classification performance on ImageNet-1K.
    }
    \vspace{1.0pt}
    \begin{tabular}{l | c  c }
    \Xhline{1.0pt}
    Methods & Parameters (M) & ImageNet-1k Top1 (\%) \\
    \hline
    \baseline{Full Fine-tuning}   & \baseline{86.57} & \baseline{82.26}  \\
    \hline \hline
    Liner       & 0.77  & 80.95  \\
    VPT         & 0.78 & 81.68 \\
    AdaptFormer-1 & 0.80 & 82.33  \\
    AdaptFormer-4 & 0.85 & 82.26  \\
    AdaptFormer-16 & 1.07 & 82.24  \\
    AdaptFormer-64 & 1.96 & 81.86  \\
    \Xhline{1.0pt} 
    \end{tabular}
    \label{tab:imagenet21k-1k}
\end{table}

\rebuttal{
We point out that in order to evaluate the adaptation performance across datasets, it's an unreasonable setting to fine-tune the ImageNet-1k dataset with the ImageNet-21k pre-trained weights. This is because \textbf{ImageNet-1K} is a subset of the \textbf{ImageNet-21K} as introduced in~\cite{ridnik-2021-imagenet21k}. In contrast, in all the previous experiments, there is no overlap between the fine-tuning and pre-trained datasets. However, we document the experiments of fine-tuning with the ImageNet-1k dataset for the completeness.}

\rebuttal{We adopt exactly identical training configurations to conduct experiments in this subsection. We experiment with \texttt{middle dimension = \{1, 4, 16, 64\}} on ImageNet-1K, and the results are shown Table~\ref{tab:imagenet21k-1k}.
}

\rebuttal{
\textbf{Results.} Comparing the results of AdaptFormer with different  \texttt{middle dimension} (\{1, 4, 16, 64\}) on ImageNet-1K, we find that AdaptFormer with the smallest number of parameters (\textbf{AdaptFormer-1}) achieves the best top-1 accuracy (82.33\%). Furthermore, when the `middle dimension` increases from 1 to 4 or 16, AdaptFormer has a slight performance drop (AdaptFormer-4 (-0.07\%) and AdaptFormer-16 (-0.09\%)). Further increasing the \texttt{middle dimension} to 64 will cause a relatively clear performance drop (-0.47\%).
}

\rebuttal{
\textbf{Discussions.} Although our AdaptFormer-64 does not have a clear advantage compared with VPT~\cite{jia-2022-vpt}, our AdaptFormer-1 outperforms VPT by +0.65\% top-1 accuracy with only 0.02M additional parameters. Besides, the trend of \textit{classification accuracy} changing with \texttt{middle dimension} on ImageNet-1k is different from other datasets in our paper, \eg, AdaptFormer with \texttt{middle dimension=64} achieves better top-1 accuracy than with \texttt{middle dimension=1} on CIFAR-100. We empirically find introducing a small number of parameters (AdaptFormer-1) is sufficient for ImageNet-1K fine-tuning, while more introduced parameters will make a larger change to the original model and make it harder for optimization since  ImageNet-1K is a subset of ImageNet-21K. However, for other datasets (e.g., CIFAR-100) with no overlap between the fine-tuning datasets and the pre-trained datasets, more introduced parameters are needed for learning better domain knowledge.
}

\subsection{\rebuttal{Extended experiments on middle dimension}}
\rebuttal{
We conduct the extended ablation studies on the middle dimension design in this sub-section. 
We aim to seek for a trade-off between model capacity (i.e., potential) and adaptation efficiency. In fact, the middle dimension has a main influence on the parameter size of adapter. The higher dimension brings more parameters while the efficiency and storage are limited. As shown in Table~\ref{tab:mid_dimension}, we evaluate several numbers of middle dimension and found that using 64 is optimal to achieve accuracy, light-weight storage, and efficiency.
}

\begin{table}[ht]
\addtolength{\tabcolsep}{-2.2pt}
\renewcommand{\arraystretch}{1.2}
    \centering
    \caption{\rebuttal{\textbf{\ourabbr ablation experiments} with ViT-B/16 on \textcolor{Video}{SSv2}. The experimental results on middle dimension are investigated.
    }}
    \vspace{1.0pt}
    \begin{tabular}{l | c  c c }
    \Xhline{1.0pt}
    Middle Dimension & Parameters (M) & SSv2 Top1 (\%) & NUS-WIDE mAP (\%) \\
    \hline 
    1 & 0.16  & 50.03 & 57.51  \\
    4 & 0.22  & 54.70 & 58.14 \\
    16 & 0.44  & 57.62 & 59.00  \\
    32 & 0.73  & 58.27 & 59.09  \\
    64 & 1.32  & 59.02 & 59.07  \\
    128 & 2.51  & 58.95 & 59.49  \\
    256 & 4.87  & 58.87 & 59.62  \\
    512 & 9.59  & 58.98 & 59.82  \\
    \Xhline{1.0pt} 
    \end{tabular}
    \label{tab:mid_dimension}
\vspace{-1em}
\end{table}

\subsection{\rebuttal{Analysis on the fine-tuning time and inference latency}}
\rebuttal{To analysis the computational efficiency, we compare the fine-tuning time and inference time on a single NVIDIA A100-40G GPU. We utilize SSv2 video classification for this part. For fine-tuing, we experiment with batchsize of 32. For inference, we test the latency with multiple batch sizes to get a comprehensive comparison under various inference scenarios. All the time is measured in milliseconds averaged  over 100 trials. The results are summarized in Table~\ref{tab:computation_1} and Table~\ref{tab:computation_2}. As shown in Table~\ref{tab:computation_1}, AdaptFormer only costs less than a half of the fine-tuning time compared with the full-tuning. Moreover, AdaptFormer significantly outperforms linear probing in terms of accuracy with a slight longer fine-tuning time. For inference, AdaptFormer introduce a negligible FLOPs and latency compared with the Linear/Full-tuning.
}
\begin{table}[ht]
\addtolength{\tabcolsep}{-2.2pt}
\renewcommand{\arraystretch}{1.2}
    \centering
    \caption{\rebuttal{\textbf{Fine-tuning time of a single forward-backward step averaged over 100 trials.}}
    }
    \vspace{1.0pt}
    \begin{tabular}{l | c   }
    \Xhline{1.0pt}
    Methods & Latency (B=32)\\
    \hline
    Full-tuning   & 355.0 ms  \\
    Linear     & 140.2 ms  \\
    VPT        & 210.3 ms \\
    AdaptFormer  & 162.2 ms   \\
    \Xhline{1.0pt} 
    \end{tabular}
    \label{tab:computation_1}
\end{table}

\begin{table}[ht]
\addtolength{\tabcolsep}{-2.2pt}
\renewcommand{\arraystretch}{1.2}
    \centering
    \caption{\textbf{Inference time of a single forward step averaged over 100 trials.}
    }
    \vspace{1.0pt}
    \begin{tabular}{l | c |c c c }
    \Xhline{1.0pt}
    Methods & Flops (B=1) & Latency (B=1)  & Latency (B=16) & Latency (B=32)\\
    Linear/Full-tuning   & 78.915G & 11.1 ms & 22.4 ms & 42.3 ms \\
    VPT   & 79.029G & 11.3 ms & 22.9 ms & 42.4 ms \\
    AdaptFormer   & 79.840G & 11.9 ms & 23.2 ms & 42.8 ms \\
    \Xhline{1.0pt} 
    \end{tabular}
    \label{tab:computation_2}
\end{table}

\subsection{\rebuttal{Discussion about ImageNet and Kinetics Pre-training}}

\rebuttal{The type of spatiotemporal attention (\textbf{divided} \textit{vs.} \textbf{joint}) determines whether the performance of the model pre-trained on ImageNet can outperform the model pre-trained on Kinetics.}

\rebuttal{A similar phenomenon has been discussed in recent work, Uniformer~\cite{li2022uniformer}, independently. We borrow the experimental results from Table 4(c) in Uniformer paper~\cite{li2022uniformer}. Specifically, the \textbf{divided} spatiotemporal attention prefers ImageNet to Kinetics-400 for the pre-training dataset. The performance of the \textbf{divided} attention model pre-trained on ImageNet outperforms the model pre-trained on Kinetics-400. On the contrary, the \textbf{joint} spatiotemporal attention prefers Kinetics-400 to ImgaeNet. The \textbf{joint} attention model attains higher top-1 accuracy with Kinetics-400 pretraining compared to ImageNet (53.8 \textit{vs.} 52.0).}

\rebuttal{We adopt the \textbf{joint} spatiotemporal attention for all video-related experiments in this work (introduced in Appendix~\ref{Sec:joint-setting}). Therefore, our experimental phenomenon is consistent with the joint attention in~\cite{li2022uniformer}, i.e., Kinetics pretraining is preferable.}

\subsection{Implementation}

\begin{algorithm}[ht]

\caption{Implementation of AdaptMLP in PyTorch-like style.}\label{alg:code}
\definecolor{codeblue}{rgb}{0.25,0.5,0.5}
\definecolor{colorred}{RGB}{197, 49, 124}
\lstset{
  backgroundcolor=\color{white},
  basicstyle=\fontsize{8.8pt}{8.8pt}\ttfamily\selectfont,
  columns=fullflexible,
  breaklines=true,
  captionpos=b,
  commentstyle=\fontsize{7.2pt}{7.2pt}\color{codeblue},
  keywordstyle=\fontsize{7.2pt}{7.2pt}\color{colorred},
}
\begin{lstlisting}[language=python]
class AdaptMLP(nn.Module):
    def __init__(self, original_mlp, in_dim, mid_dim, dropout=0.0, s=0.1):
        super().__init__()
        self.original_mlp = original_mlp  # original MLP block
        # down --> non linear --> up
        self.down_proj = nn.Linear(in_dim, mid_dim)
        self.act = nn.ReLU()
        self.up_proj = nn.Linear(mid_dim, in_dim)
        self.dropout = nn.Dropout(dropout)
        self.scale = s  # scaling factor
        # initialization
        nn.init.kaiming_uniform_(self.down_proj.weight)
        nn.init.zeros_(self.up_proj.weight)
        nn.init.zeros_(self.down_proj.bias)
        nn.init.zeros_(self.up_proj.bias)
        # freeze original MLP
        for _, p in self.original_mlp.named_parameters():
            p.requires_grad = False

    def forward(self, x):
        down = self.down_proj(x)
        down = self.act(down)
        down = self.dropout(down)
        up = self.up_proj(down)
        output = self.original_mlp(x) + up * self.scale
        return output
        
\end{lstlisting}
\vspace{-2.5mm}
\end{algorithm}

The core part of \ourabbr is replacing the original MLP with AdaptMLP, which consists of the frozen original MLP and newly introduced \texttt{Down}~$\rightarrow$~\texttt{ReLU}~$\rightarrow$~\texttt{Up} layers, which are tunable at the fine-tuning stage.   
Algorithms~\ref{alg:code} provides the implementation of AdaptMLP written in PyTorch~\cite{NEURIPS2019_9015}.

For more implementation details, please refer to the provided source code.

\end{document}